*Article*

# Hybrid Artificial Intelligence Strategies for Drone Navigation

Rubén San-Segundo [1,*] , Lucía Angulo [1], Manuel Gil-Martín [1,*] , David Carramiñana [2]  and Ana M. Bernardos [2]

1    Speech Technology and Machine Learning Group, Information Processing and Telecommunications Center, Universidad Politécnica de Madrid, ETSI Telecomunicación, Av. Complutense, 30, 28040 Madrid, Spain; lucia.anguloc@alumnos.upm.es

2    Simulation and Data Processing Group, Information Processing and Telecommunications Center, Universidad Politécnica de Madrid, ETSI Telecomunicación, Av. Complutense, 30, 28040 Madrid, Spain; d.carraminana@upm.es (D.C.); anamaria.bernardos@upm.es (A.M.B.)

\*    Correspondence: ruben.sansegundo@upm.es (R.S.-S.); manuel.gilmartin@upm.es (M.G.-M.); Tel.: +34-910672225 (R.S.-S.)

**Abstract:** Objective: This paper describes the development of hybrid artificial intelligence strategies for drone navigation. Methods: The navigation module combines a deep learning model with a rule-based engine depending on the agent state. The deep learning model has been trained using reinforcement learning. The rule-based engine uses expert knowledge to deal with specific situations. The navigation module incorporates several strategies to explain the drone decision based on its observation space, and different mechanisms for including human decisions in the navigation process. Finally, this paper proposes an evaluation methodology based on defining several scenarios and analyzing the performance of the different strategies according to metrics adapted to each scenario. Results: Two main navigation problems have been studied. For the first scenario (reaching known targets), it has been possible to obtain a 90% task completion rate, reducing significantly the number of collisions thanks to the rule-based engine. For the second scenario, it has been possible to reduce 20% of the time required to locate all the targets using the reinforcement learning model. Conclusions: Reinforcement learning is a very good strategy to learn policies for drone navigation, but in critical situations, it is necessary to complement it with a rule-based module to increase task success rate.

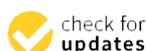

## 1. Introduction

Nowadays, the use of Unmanned Aerial Vehicles (UAVs) or drones has increased significantly in many domains, including military and civil applications [1,2]. Civil applications include aerial surveillance, parcel delivery, precision agriculture, intelligent transportation, search and rescue operations, post-disaster operations, wildfire management, remote sensing, and traffic monitoring [3]. These developments have been possible thanks to a fast deployment of radio communication interfaces, sensors, device miniaturization, global positioning systems (GPSs), and artificial intelligence (AI) technologies. Using drones has important advantages such as cost effectiveness, fast mobility, and easy deployment [4], but also must face relevant challenges like scalability or computer power consumption [5].

Artificial intelligence and machine learning (ML) algorithms are having an important development [6] and a significant impact across a great variability of sectors ranging from healthcare [7], smart cities [8], natural language processing and human–computer interaction [9], to transportation and logistics [10]. Similarly, these algorithms have an important role in drone applications because ML enhances drones' capabilities in navigation [11], object detection [12], or mission planning [13].

This paper is focused on the development and evaluation of hybrid artificial intelligence strategies for drone navigation in simulated environments. The navigation problem can be defined as follows: "given a starting and a goal point or a set of goal points defined in the same frame of reference, a system should use prior knowledge if available or accumulated knowledge to plan and execute a feasible trajectory from a start to a goal configuration" [14]. Navigation capability includes several abilities like perception, localization, motion planning, and motion control. Additionally, in some cases, it is also necessary to deal with obstacle avoidance [14].

UAVs can navigate complex environments where finding a collision-free path is essential to avoid losing the drone. Some traditional and widely used algorithms are the Dijkstra algorithm [15], A-star algorithm [16], and random tree (RRT) algorithm [17]. Finding the optimal path requires ensuring the optimal use of energy and maximizing the efficiency of mission execution. In the literature, most previous works address the path planning problem of SUAVs as an optimization problem [18]. In many of these studies, the algorithms work well when the whole environment information is available for making the decision, but in many cases (especially when dealing with moving obstacles), this information is not available, and a local decision must be taken based on the observation space.

Reinforcement learning (RL) is a branch of machine learning methods that involves training an agent based on interactions with an environment [19]. The agent learns to maximize a numerical reward received from the environment after their actions. This iterative process allows the agent to learn the optimal policy for achieving its goal [20]. One of the main advantages of RL is that it does not need a prefixed training set, and the agent will adjust the action based on continuous feedback to maximize the final reward, being able to adapt its behavior to changing environments. RL is being used to learn how to play different games [21]. In some previous works, different RL algorithms for drone navigation were used: the Q-learning algorithm [22], Proximal Policy Optimization (PPO) algorithm [23], or Double Deep Q-Network DDQN (DDQN) [24]. In this work, the PPO algorithm has been used because it is one of the most used algorithms. PPO provides a very good compromise between training time and performance and provides good stability during the training process. PPO optimizes an agent policy by maximizing an objective function that compares the new policy to an old policy, ensuring that the update is not too large, which helps in maintaining stable training.

In many of these previous works, the researchers considered navigation problems (in two or three dimensions) in simulated environments where many episodes over different scenarios can be executed to make the agent (drone) learn what to do in different situations. RL methods have demonstrated significant advantages for autonomous drones in self-navigation, particularly when relevant information is missing. Through trial-and-error interactions with the environment, RL enables drones to develop navigation policies that account for missing data. RL's adaptability ensures effective navigation in dynamic and unpredictable scenarios. These advantages allow using RL algorithms to learn from experience and adapt to dynamic environments [25]. RL provides a powerful solution for autonomous drones to overcome challenges like incomplete or missing information and achieve successful self-navigation [26,27].

To develop a robust drone navigation system, it is interesting to complement the knowledge learnt through machine learning algorithms with rules developed by expert humans. These rules increase the reliability of the system when dealing, for example, with complex scenarios involving obstacles and/or moving targets. This paper describes the development of a drone navigation system based on hybrid AI combining RL and a rule-based engine for defining the drone navigation policy. The main contributions of this paper are the following:

- This paper describes the development of hybrid artificial intelligence strategies for drone navigation in simulated environments.
- The hybrid AI combines deep learning models obtained using reinforcement learning with human rules developed by experts. These expert rules increase the robustness of the drone navigation strategy.

- The system incorporates explainability strategies to better understand the drone's decision according to the available information (local observation space).
- The system also incorporates human interaction strategies, allowing the combination of automatic and supervisor's decisions during the navigation process.
- This paper also proposes an evaluation methodology that includes the simulation of several scenarios, and the computation of performance metrics adapted to each scenario.

## 2. System Description and Navigation Tasks

The next figure (Figure 1) represents a diagram of the multiagent system, including the main modules.

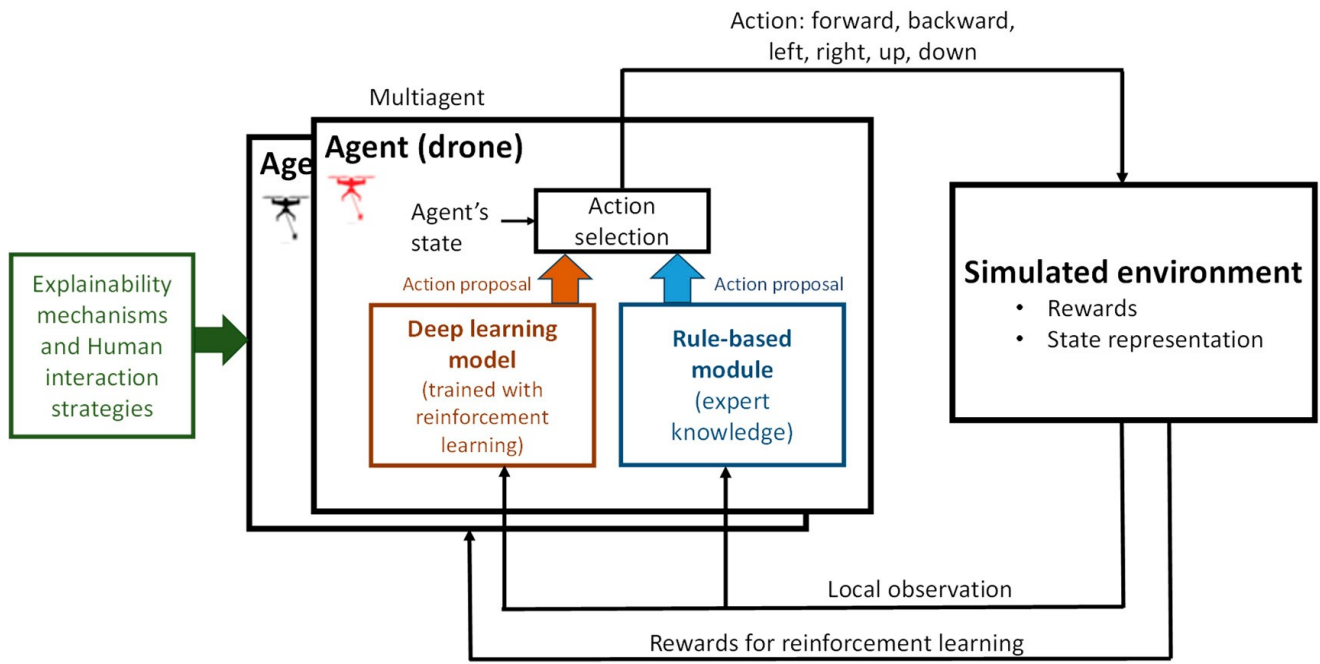

**Figure 1.** Diagram of the global system.

The global system is composed of two main modules:

- The agent (drone) module generates the next drone action using the local observation of the agent. This process selects the drone's action proposed from the rule-based module or the deep learning model, depending on the agent state. The input of this module is the local observation space (a 20×20 submatrix, explained in Section 3.1). The output is the next drone's action. The agent can perform six possible actions in a 3D environment (forward, backward, left, right, up, down) or four in a 2D environment (forward, backward, left, right). The agents/drones perform their action in sequence considering a specific order. When all the agents have executed a new action, the system has completed one cycle.
- The simulated environment is a global representation of the scene including drones, targets, and obstacles. The environment includes two agents (drones) that must complete different drone navigation tasks considering a configurable number of targets (in green circles) and avoiding several obstacles (represented with bombs in black). An example is shown in Figure 2. Every object includes a Z coordinate in black, indicating its height. This way, it is possible to simulate 2D and 3D navigation scenarios. The inputs to this module are the drones' actions, and the outputs are the new environment state and the reward associated with every action. This reward is used only during training.
- Additionally, the system integrates several explainability mechanisms and human interaction strategies (green module). The explainability mechanisms are more focused on helping with the algorithm development while the human interaction strategies are oriented to allow the intervention of a human operator. These tools allow the representation of internal information regarding the drones and the environment.

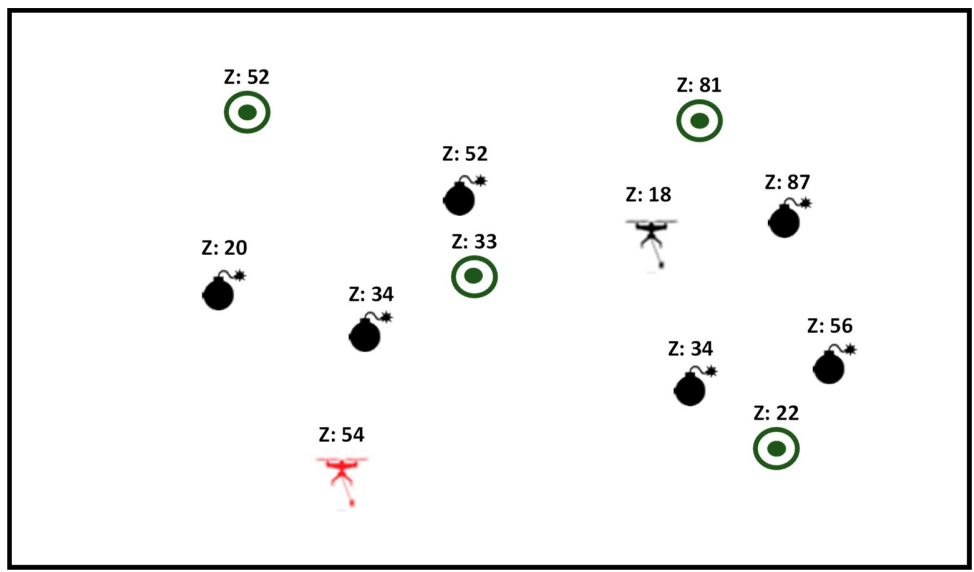

**Figure 2.** Visual representation of the multiagent environment.

In this work, two navigation tasks (reaching located targets and searching for new targets) have been considered with several scenarios per task. The objective of analyzing several scenarios per task is to have a more complete analysis covering a wider range of applications.

### 2.1. Reaching Located Targets

In this task, the drones must navigate through an environment to reach all the targets. The simulated environment is a global representation of the scene including drones, targets, and obstacles The drones know the target coordinates, but they do not know the whole environment; drones only see a local observation of the environment. The scenarios evaluated in this task are as follows:

- Considering static or moving targets without obstacles. In the case of moving targets, they try to reach the bottom of the screen to escape from the drones.
- Static targets including obstacles. Several obstacles are included, and the drones must avoid them to reach the targets.
- Considering a 3D navigation problem with static targets and a different number of obstacles. A third dimension increases the flexibility of the drone's movements.

### 2.2. Searching for Targets

In this task, the drones must search the environment looking for targets. The drones do not know the targets' coordinates and they only see a local observation space. The objective is to find all the targets as soon as possible, with the lowest number of movements or cycles.

The targets are distributed randomly in the game scene but using specific patterns. Considering these patterns, it is possible to train specific search strategies to reduce the time to discover all the targets. The targets are organized in groups randomly distributed in the scene. Every group occupies a small zone, and the targets are also randomly distributed inside the group zone. These small areas are defined as squares with a size of 20% of the total width and height. The analyzed scenarios are as follows:

- Considering static or moving targets. In the case of moving targets, when one target is detected, the rest try to reach the bottom of the screen to escape from the drones.

- Including obstacles. Several obstacles are included, and the drones must avoid them during the search.

## 3. Materials and Methods

This section describes in detail the algorithms implemented in the module of the drone navigation system. Additionally, a description of the data used for training and evaluating the different proposals is included.

### 3.1. Simulated Environment

The simulated environment has been developed based on the PettingZoo environment definition [28]. PettingZoo is a simple interface, developed in Python, able to represent general multi-agent reinforcement learning problems. PettingZoo includes many environments as examples, and tools for designing custom environments.

The 2D environment state space is represented as a $200 \times 200$ matrix with different values per cell: $-1.0$ values are assigned to forbidden areas (obstacles and game margins to avoid collisions) (Figure 3). In the case of a 3D environment, the environment state space is a 3D matrix.

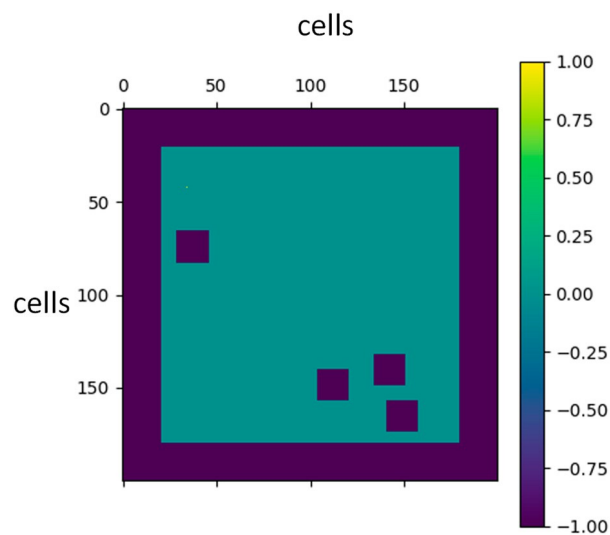

**Figure 3.** Environment state representation: a 200 × 200 matrix of the whole environment.

### Reward strategy

The reward strategy includes the following two main aspects:

- Target reached by an agent or drone: when a target is reached by an agent, the corresponding agent's reward is increased by 1, emphasizing the importance of reaching a target.
- Partial rewards depending on the distance to the target: An additional reward based on the agent's proximity to the target has been added. This reward is associated with the distance reduction between the drone and the target. Once the drone executes a movement, its reward increases according to the distance (to target) reduction. It is important to remark that these partial rewards are only applicable when the drone knows the target coordinates. These partial rewards introduce a reward-shaping mechanism that encourages agents to reduce the distance to their target. Including these partial rewards allows significantly reducing the training time.

### Distance to the target

The following two alternatives have been analyzed to compute the distance to the target:

- The first one consists of computing the minimum Euclidean distance between two points in the environment matrix. This distance generates frequent problems when dealing with obstacles. During training, the agent learns to take the shortest path to reach the target, considering a straight line. If there is an obstacle in the middle, the agent can hit it, or at least the obstacle is detected and surrounded. If the agent detects the obstacle and goes backward, it gets stuck; thus, the learnt policy forces the agent to go in a straight line toward the target, but when the obstacle is detected, the agent goes backward.

- The second alternative consists of redefining the distance between the agent and the target to allow several possible paths with the same distance, not only one. The way to consider several paths with the same distance is by computing the distance by summing vertical and horizontal movements uniquely, as shown in Figure 4. As shown in this figure (right side), there are several paths with the same distance. This solution increases the agent's flexibility to avoid obstacles. But this flexibility was not enough, and it did not solve all the cases to avoid obstacles; when the drone movement must be horizontal or vertical, the flexibility disappears. Because of this, it was necessary to implement a specific strategy (based on rules) to avoid obstacles (Section 3.2.1).

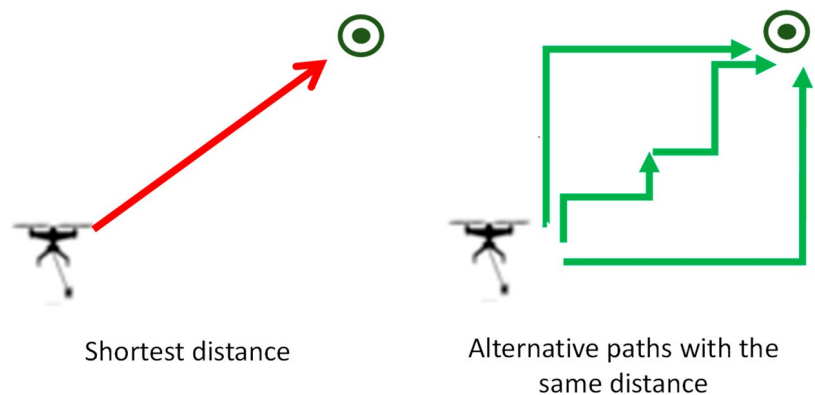

Shortest distance                          Alternative paths with the same distance

**Figure 4.** Alternative paths with the same distance.

## 3.2. Agent Module

The main target of this module is to define the next action according to the drone navigation policy. The next drone action is selected from the rule-based module or from the deep learning model depending on the drone state and the local observation of this drone. The local observation of a drone is represented by a 20 × 20 submatrix of the state space (the number of 2D or 3D dimensions depends on the scenario), centered in the agent (Figure 5). Every cell in this matrix includes the reward that the agent would obtain if moving to this point: −1.0 values are assigned to forbidden areas (margins and obstacles). In this figure, the center is marked with a −1.0 (only for representation).

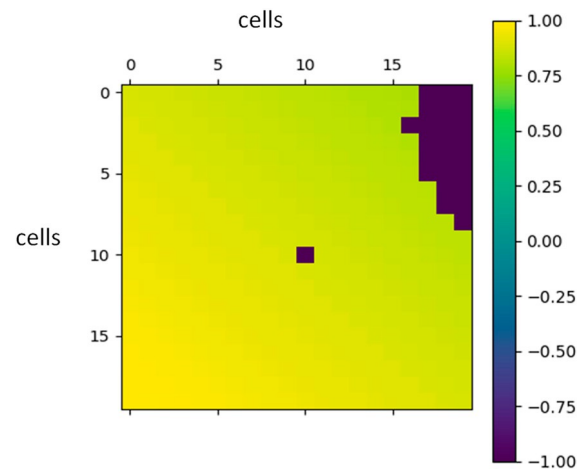

**Figure 5.** Observation space representation: a 20 × 20 submatrix.

The next sections describe the different hybrid strategies depending on the navigation task.

### 3.2.1. Task 1: Reaching Located Targets

In this task, the drones must navigate to the different targets until reaching all of them. Every drone is assigned a specific target that it must reach (the closest target). When a target is reached, it is marked as seen and the drone is reassigned a new target (the next closest one). It is important to comment that when there are more unseen targets than agents (drone), every drone has different assigned targets to parallelize the task resolution.

For drone navigation in this task, there are two main agent states: the normal navigation state (first state) and obstacle detected in the path to the target (second state).

**First state: normal navigation**

In the normal navigation state, the policy is defined by a deep learning module trained using a reinforcement learning (RL) algorithm. Stable-Baselines3 [29] is the RL toolkit used in this work because of the large amount of available RL algorithms and its flexibility in the integrations process. After an initial analysis of the different RL algorithms, PPO [30] was chosen because it provides a very good compromise between training time and performance and offers a good stability during the training process. PPO iteratively improves the agent's policy to maximize cumulative rewards. Its main advantage lies in balancing exploration and exploitation by judiciously constraining policy updates.

In this paper, the PPO-Clip version has been used from the Stable-Baselines3 [29] toolkit. This library implements PPO-Clip with a clip range parameter equal to 0.3. The clipping parameter is a function of the current progress remaining (a value between 0 and 1). This parameter was set to 0.3 following the suggestion of the Stable-Baseline3 library. This value is a good compromise for a wide range of applications. Additionally, several prelaminar experiments were conducted while modifying this parameter, and the best results were obtained with this proposed value. PPO-Clip performs better than PPO-Penalty because in this implementation, the new policy would not update so far away from the old policy, avoiding oscillations during the learning process.

The policy is implemented based on a deep learning architecture where the input is the local observation space (representation of the context observed by the agent) and the output is an array of probabilities for all the possible actions or movements of the agent: forward, backward, left, right, up, and down (6 possible actions).

This deep learning architecture is composed of two main parts (Figure 6):

- A feature extractor for extracting features from high-dimensional observations. This module includes two CNN layers with 32 kernels and ReLU functions.
- A fully connected network that maps the features to actions, including two fully connected layers with 128 neurons each and a final layer with P outputs and the Softmax function (classification). The number of outputs is 6 (forward, backward, left, right, up, down) or 4 for 2D scenarios (forward, backward, left, right).

**Second state: avoiding an obstacle in the path to the target**

Although a new distance was defined to provide certain flexibility to the drone path, there are situations where the drone cannot find an alternative path without obstacles. These cases are automatically detected by the drone when the drone repeats several movements in a cycle without decreasing the distance to the target. In this circumstance, the drone understands that it was not able to find an alternative path without obstacles and starts executing the avoiding strategy based on rules. In this state, the agent is forced to go around the obstacle, describing a circumference as shown in Figure 7. This behavior is provoked by changing the agent target; the system generates a sequence of fictitious targets that go around the obstacle.

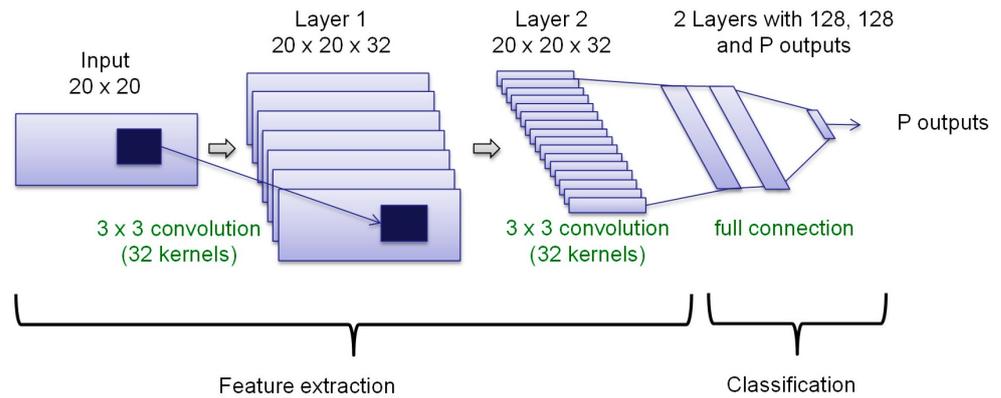

**Figure 6.** Deep learning architecture for the policy model.

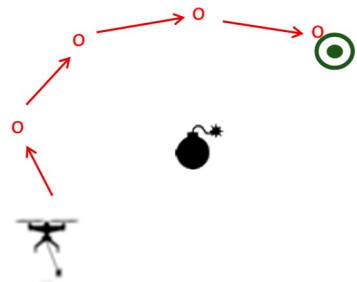

**Figure 7.** Path followed by the drone to avoid the obstacle thanks to the rule-based module.

The process starts when an obstacle is detected between the agent and the target, and the drone is not able to find an alternative path without obstacles (Figure 8). In this situation, the system modifies the agent's target, situating a fictitious target to a certain angle with respect to the line connecting the agent and target. Once the agent reaches the fictitious target, a new real target is assigned (in the circumference centered in the obstacle), and the process starts again. This process can be repeated several times until the agent is able to find a free path to the real target. The next figure shows an example of the path followed by the drone to avoid obstacles by reaching intermediate fictitious targets.

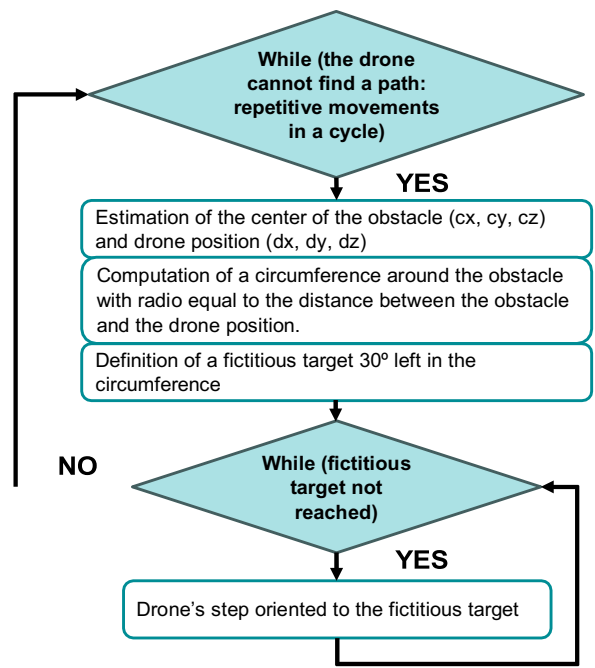

**Figure 8.** Flowchart for avoiding an obstacle in the path to the target.

As shown, the algorithm is continuously (at each step) evaluating the existence of an obstacle between the drone and the target. If the target or the obstacle changed their position, the algorithm was able to adapt itself to this new situation.

### 3.2.2. Task 2: Searching for Targets

In this scenario, the drones must search the environment looking for targets. The drones do not know the targets' coordinates and they only see a local observation space. The objective is to find all the targets as soon as possible, with the lowest number of movements or cycles.

The targets are distributed randomly in the game scene but follow specific patterns. Considering these patterns, it is possible to train specific search strategies to reduce the time to discover all the targets. In this work, the targets are organized in groups randomly distributed in the scene. Every group occupies a small zone, and the targets are also randomly distributed inside the group zone.

For drone navigation in this task, the system considers three main agent states: exhaustive search (state 0), local search around the last target detected (state 1), and obstacle detected during local search (state 1.1).

**First state: exhaustive search**

All the agents (drones) start doing an exhaustive search. The navigation is guided by a rule-based module. During the exhaustive search, the drone movements are defined by a set of expert rules (Figure 9), trying to cover the whole environment in vertical paths like in Figure 10.

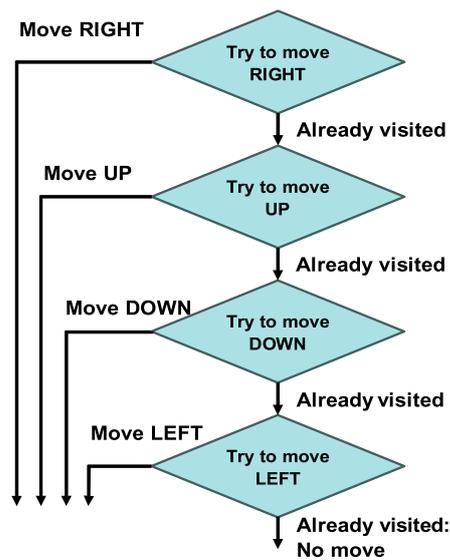

**Figure 9.** Flowchart for exhaustive search: drone movement decision. Priority of movements.

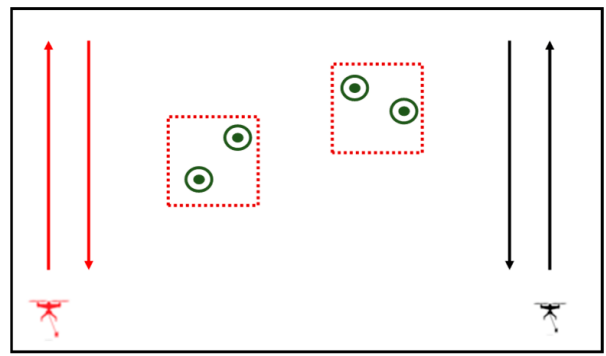

**Figure 10.** Exhaustive search in vertical paths (arrows indicate drone's direction).

**Second state: local search around the last target detected**

When one target is detected, the drone stops the exhaustive search, and a reinforcement learning model guides the drone movements to search in the local area, close to the first discovered target. This model has been trained according to the pattern commented above: the targets are organized in several groups. The main aspects while training the model are the following:

- For the drone observation space, the same observation space used for the previous scenario (reaching targets) has been considered: a 20×20 submatrix of the state space, centered in the agent, including for each cell the reward that the agent would obtain moving to this point (−1.0 values are assigned to forbidden areas, scene, and object margins). In this case, the first detected target is the reference to compute the agent_target distances. In the case of having several detected targets in the same zone, the reference for computing the distance is the average target position.
- For the reward strategy, the system computes a distance based on the vertical and horizontal movements between the current position of the drone and the reference, defined as the average position along the already discovered target.

When one target is detected by a drone, only this drone focuses on this local zone while the rest maintain an exhaustive search for dealing with other groups of targets. When a drone spends some time looking in a local zone without success, it changes its behavior to an exhaustive search again (state 0). This change is necessary to provide the possibility to search for several groups with the same drone.

**Third state: avoiding an obstacle in the path to the target during the local search**

Finally, it is important to comment that the same rule-based strategies for avoiding obstacles during the local search are also integrated in this case.

*3.3. Combination of Deep Learning and Rule-Based Policies*

This section summarizes the combination of deep learning and rule-based policies based on the navigation task. This combination is based on states depending on the following tasks:

- Task 1: reaching located targets:
  - First state: normal navigation using the deep learning policy learnt using reinforcement learning.
  - Second state: Avoiding an obstacle in the path to the target. The rule-based engine is used to define fictious targets that allow avoiding the obstacle. The drone is maintained in the second state while it cannot find a direct path to the target. This situation is detected when the drone performs repetitive movements in sequence.
- Task 2: searching for targets:
  - First state: Exhaustive search. Using the rule-based engine for covering the whole area.
  - Second state: Local search around the last target detected. When the first target is found, the drone changes to the second state and the deep learning architecture is used for navigation. The target is to find close targets in the local area of the first detected target. After a certain time without finding new targets, the drone returns to the first state.
  - Third state: Avoiding an obstacle in the path to the target during the local search. Like task 1, the rule-based engine for obstacle avoidance is used when the drone cannot find a direct path to the target.

*3.4. Explainability Mechanisms*

In the system, two main explainability mechanisms have been integrated to help with the algorithm development: LIME (Local Interpretable Model-Agnostic Explanations)[31] and SHAP (SHapley Additive exPlanations)[32].

### 3.4.1. Local Interpretable Model-Agnostic Explanations (LIME)

LIME is a technique that approximates any black box machine learning model with a local interpretable model to explain each individual prediction, providing an explanation of the decisions made by an agent in certain situations.

Figure 11 shows the local observation, a matrix where the agent's observation space is represented. This figure shows two obstacles corresponding to the black blocks in the next image (bottom and right parts).

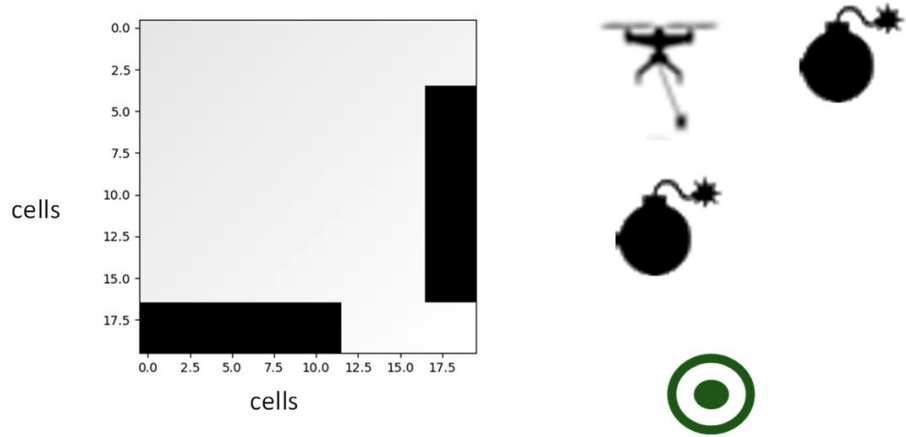

**Figure 11.** Local observation.

After applying LIME, it is possible to see the contribution of each cell or pixel in the local observation for each action. LIME uses colors in its representations to indicate the importance of different regions of the image in relation to the model prediction (red and green):

- Green: The areas highlighted in green indicate regions of the image that have contributed positively to the predicted classification. In other words, they are the parts of the image that have led the neural network to predict the specific class.
- Red: The areas highlighted in red, on the other hand, indicate regions that have had a negative impact on the classification. These are regions that provide negative information related to this action.

These colors are used to provide an intuitive visual representation of how different parts of the image affect the model output in terms of classification probability. The intensity of the color indicates the magnitude of the contribution of each region.

Figure 12 represents the contribution of each pixel/cell in the local observation space for each action. In this case, there are four actions in a 2D environment: 1 (forward), 2 (backward), 3 (left), and 4 (right). As shown, the obstacles in the left and bottom part are suggesting that the best actions (more green pixels with positive impacts) are 1 and 3.

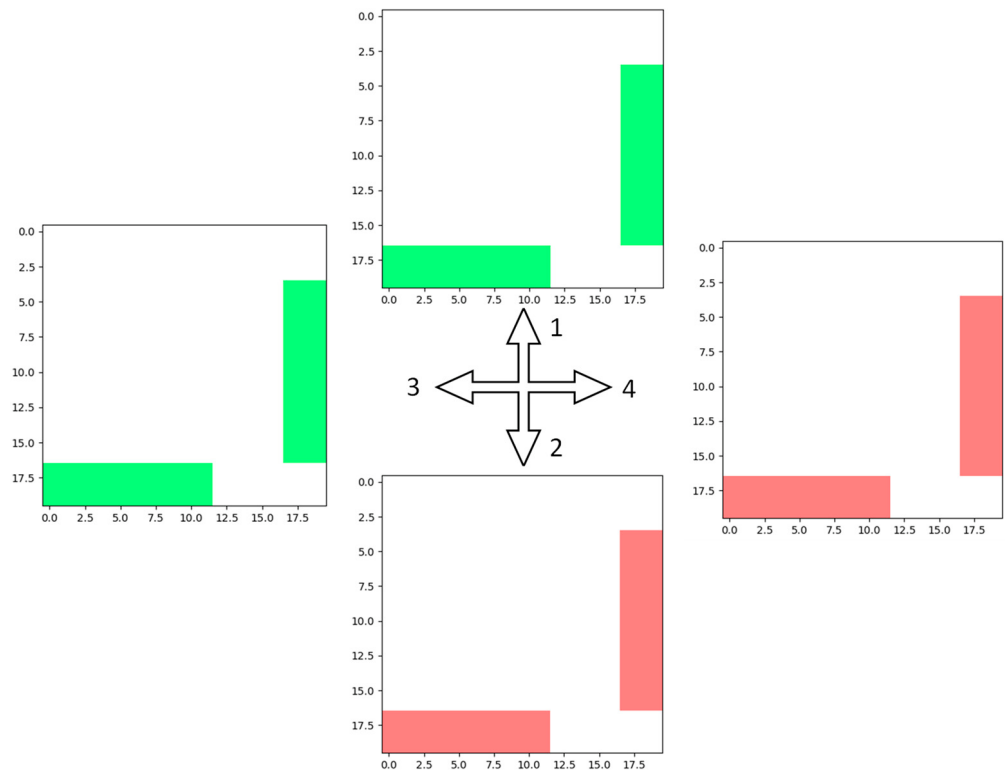

**Figure 12.** Analyses of actions 1 (forward), 2 (backward), 3 (left), 4 (right) in a 2D environment.

### 3.4.2. SHapley Additive exPlanations (SHAP)

Additionally, the study also includes an analysis using SHAP to obtain more details about the explainability of the decisions made by the agents/drones. This method is based on game theory and uses Shapley values to compute contributions to each feature in the prediction of a model. These features are pixels when analyzing an image.

SHAP explanations are often represented graphically using bar charts or scatter plots, where each bar or point represents the contribution of a specific feature to a particular prediction. In SHAP, the areas appearing in pink and blue on the graphs indicate the impact of the contribution of each pixel to the model prediction value. The interpretation of the colors in SHAP graphs is as follows:

- Blue: negative contribution (low): The blue areas represent features that contribute negatively to the model's prediction value. In a classification context, this could be interpreted as features that decrease the probability of the predicted class.
- Pink: positive contribution (high): The pink areas represent features that contribute positively to the model's prediction value. In a classification context, this could be interpreted as features that increase the probability of the predicted class.
- Color intensity: magnitude of contribution. The color intensity (whether blue or pink) indicates the magnitude of the contribution of that specific feature. The darker the color, the greater the magnitude of the contribution.

Figure 13 shows an example of a SHAP analysis with the actions sorted from more to less probability (left to right). The action with the highest positive contribution is action number 1 (move up), and the action with the highest negative contribution is action number 2 (move down).

LIME and SHAP mechanisms have been used to evaluate the relevance of each point or pixel in the local observation space when selecting the drone action by the deep learning model. These mechanisms have provided information about the contribution of each pixel (positive or negative) and the intensity for each possible drone's action. For example, when detecting an obstacle ahead, the pixels showing the obstacle have a negative contribution for going in the same direction, but a positive contribution for the opposite action. Thanks to these mechanisms, it is possible to supervise the iterative learning process; their analyses are used to finetune the main hyperparameters involved in the reinforcement learning process (learning rate, batch size, etc.).

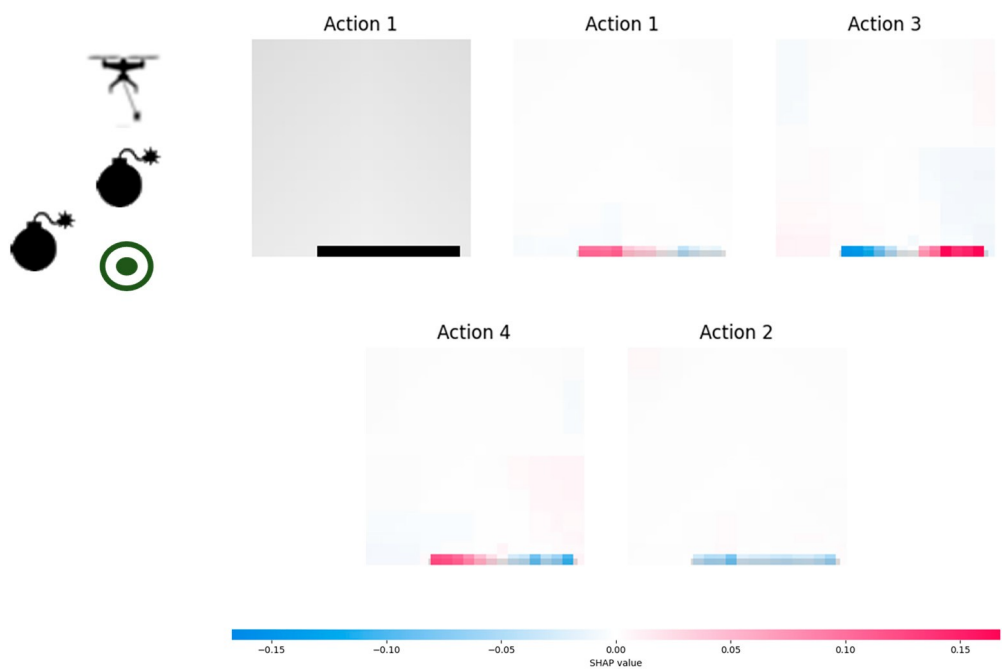

**Figure 13.** SHAP results: selected action (action 1, up) and the analyses for all actions (1 up, 2 down, 3 left, 4 right in a 2D environment).

### 3.5. Human Interaction Strategies

This section describes the functionalities incorporated to allow the human supervisor to interact with the agent, modifying its behavior. Additionally, some functionalities that help to understand the behavior of the agents (commented in the previous section) have also been incorporated. Most of the functionalities are only available when the supervisor pauses the navigation. These features are the following:

- Save the current state of the game: The current state of the game (agents, targets, and obstacles) can be saved at any moment. This current state also contains information about various aspects of the game, such as the score, the number of agents of each type, agent selection, dead agents, and rewards, among others. Several states can be stored in sequence.
- Move forward and backward in the stored state sequence; in this case, there is the possibility to go backward or forward along the state sequence.
- Advance step by step the movement of the agents.
- Perform explainability analyses for the observations of the agents: LIME and SHAP.
- Move the agents manually: change the position (Figure 14) and the target (right side) of the agents (Figure 15).
- Information window. The following image shows the visualization of the information window that shows the coordinates of each agent. These coordinates can be modified manually. Figure 16, at the bottom, shows the observation space of each agent and the explainability associated with each action (by pressing the LIME and SHAP buttons).

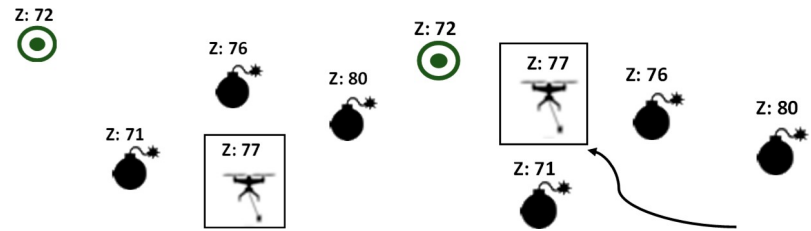

**Figure 14.** Visualization of manual movement of agents and manual change of agent's target (right button of the mouse).

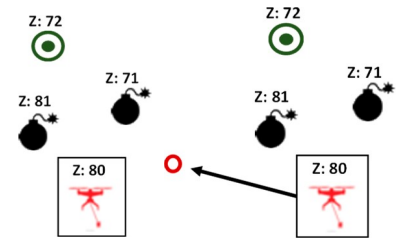

**Figure 15.** Manual change of agent's target (right button of the mouse).

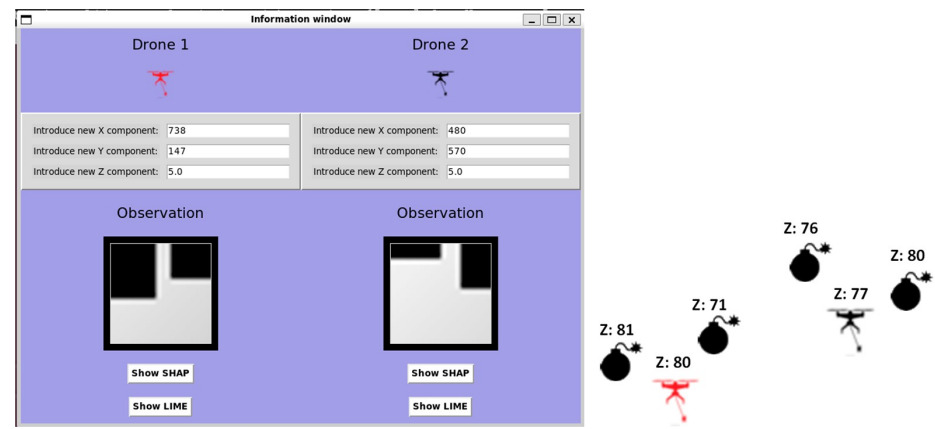

**Figure 16.** Visualization of the information window and drone positions associated with the visualization window.

Human interaction strategies have been used for recreating specific situations that can be debugged using the explainability mechanisms described above. Human interaction strategies have been used in combination with LIME and SHAP. For example, it is possible to situate the drone between two obstacles, release the drone, and analyze the drone's behavior based on its observation space (using LIME or SHAP).

In real applications, these functionalities can help the drones to solve the task, modifying their position or the assigned target. In the experiments carried out in this work, all the evaluations have been completed automatically, without human intervention.

### 3.6. Dataset

All reinforcement learning models developed in this work have been trained using specific data. The training dataset is composed of 900,000 episodes (or games) generated automatically, including more than 18 million drone movements or cycles. The system can generate different games considering the different initial positions of the drones, targets, and enemies. These positions are generated randomly, covering a wide range of possibilities.

For evaluating all the strategies in different scenarios, new games have been generated. The system has evaluated the hybrid strategies in every new scenario with 200 new games.

## 4. Evaluation Results and Discussions

This section describes the general evaluation methodology, and the analyses carried out for both tasks: reaching several located targets and searching for targets. It is important to remark that human interaction strategies (described in the previous section) have not been used during the system evaluation and testing. All the experiments have been carried out using the system automatically, and no human intervention has been considered.

### 4.1. Evaluation Methodology

The evaluation methodology is composed of two main phases. In the first phase, the reinforcement learning model is trained simulating many episodes in every task, considering several scenarios (Section 2). Every episode is defined by a setup of drones, targets, and obstacles organized in the scene.

In the second phase, the proposed system is evaluated by analyzing its behavior when dealing with new episodes generated automatically (different from those seen during the training phase). These episodes are generated for both tasks and different scenarios. The objective of considering several scenarios per task is to have a more complete analysis covering a wider range of possible applications. For each task, different evaluation metrics reporting results in different situations have been used to evaluate the navigation strategies. In both scenarios, there are quality and time-processing metrics. The quality metrics provide information about the task success (percentage of episodes completely solved by the drones) and the incidence of several possible problems like obstacle collisions (times a drone hit an obstacle, for example). To complete the analysis, time-processing metrics like the number of cycles or steps carried out by the drones for solving the task have been considered. The reason for this analysis is because in real application, drones have a limited autonomy and it is necessary to optimize their movements; it is important to solve the task, but while using as few movements as possible.

### 4.2. Task 1: Reaching Several Located Targets

In this task, the following metrics have been considered:

- Quality metrics: to analyze the task success and the incidence of possible problems.

    ◦ Percentage of episodes where all the targets were reached (task completion rate).
    ◦ Percentage of episodes where at least one agent hit an obstacle.
    ◦ Percentage of episodes where all agents hit an obstacle.

- Time-processing metrics: to analyze the drone movements required to complete the task.

    ◦ Number of cycles used for training. One cycle is completed when all the agents/drones have executed a new action. These metrics allow evaluating the amount of training required to train a good model.
    ◦ Maximum number of cycles per episode during testing. This limit simulates the situation of having a limited drone autonomy.

#### 4.2.1. Static or Moving Targets Without Obstacles

This section includes the results associated with the initial situation where the agents (drones) must reach the targets without obstacles. The baseline system consists of considering a simple reward strategy: the agents get a reward of 1.0, only when they reach a target. The rest of the movements do not provide any positive reward.

The next experiment includes partial rewards for each agent's movement (based on the distance reduction, as commented before). In this case, two situations have been evaluated: static targets and moving targets trying to reach the bottom of the screen. Table 1 includes the main results obtained when reaching targets without obstacles. The main parameters of the experimentation setup are as follows:

- Reinforcement learning algorithm: PPO with a learning rate adjusted in preliminary experiments: 0.0003.

- 2D environment.
- Two agents (drones).
- Reward = 1 × T (T is the number of targets reached) − ΔD (distance to target variation).
- Four targets randomly distributed.

**Table 1.** Main results when reaching known targets without obstacles, considering a 2D scenario.

| | Total Cycles Used in Training | Maximum Number of Cycles Per Episode in Testing | % of Episodes Reaching All Targets (Task Success) |
|---|---|---|---|
| Baseline system (Reward = T) | $18 \times 10^6$ | 200 | 57 |
| Baseline system (Reward = T − ΔD) | $6 \times 10^6$ | 200 | 100 |
| Baseline system with moving targets (Reward = T − ΔD) | $6 \times 10^6$ | 200 | 92 |

From these experiments, an important conclusion is that by including partial rewards, a better policy model can be obtained with less training effort (1/3: from 18 to $6·10^6$ cycles) than the baseline system, reaching 100% of episodes where all targets were reached. When including moving targets, there are several episodes where a target can reach the bottom of the screen without being reached and the task is not completed.

### 4.2.2. Static Targets Including Obstacles in Task 1

When an agent hits an obstacle, this agent dies and disappears from the scene. The baseline for these experiments is the best system developed without obstacles, evaluated in the previous section but including the obstacles. After that, the following several modifications were considered:

- Modifying the reward strategy: in the next experiments, the reward strategy includes a negative reward when an agent hits an obstacle.
- Modifying the agent–target distance to have alternative paths with the same distance.
- Modifying the reward strategy (training) and target correction (testing): the rule-based engine controls that the agent does not hit any obstacle introducing a correction strategy based on intermediate fictitious targets to surround the obstacle, as commented previously.

Table 2 includes the main results obtained when reaching targets with obstacles in the game. The main parameters of the experimentation setup are the following:

- Reinforcement learning algorithm: PPO with a learning rate adjusted in preliminary experiments: 0.0003.
- 2D environment.
- Two agents (drones).
- Reward = 1 × T (number of targets reached) − 1 × O (number of obstacles hit) − ΔD (distance to target variation).
- Four static targets are randomly distributed.
- Four obstacles randomly distributed.
- Total cycles used in training: $8 \times 10^6$.

The main conclusions from these experiments are the following:

- When including the obstacles, it is necessary to increase the number of cycles in training to better learn the agent policy (compared to Table 1).
- Including a negative reward when hitting an obstacle is crucial to reduce the number of situations where one drone hits an obstacle (second row).
- When considering alternative paths with the same distance, it is possible to increase the task success (percentage of episodes where all the targets are reached) from 35% to 57%.
- The only way to guarantee that the agents do not hit any obstacle is by including expert rules to detect stuck situations and surround the obstacle. In this case, the task success (percentage of cases reaching all the targets) increases to 70%, but there still are situations where the agents get stuck between several obstacles and cannot reach all the targets in a limited number of cycles (200 in these experiments).

**Table 2.** Main results when reaching known targets with obstacles, considering a 2D scenario.

| | Maximum Number of Cycles Per Episode in Testing | % of Cycles Using Deep Learning (DL) or Rule-Based (RB) Policies | % of Episodes Reaching All Targets (Task Success) | % of Episodes with (at Least) One Drone Hits an Obstacle | % of Episodes Where All Agents Hit an Obstacle |
|---|---|---|---|---|---|
| Baseline system (Reward = T − ΔD) | 200 | DL: 100 | 15 | 100 | 80 |
| Reward strategy (Reward = T − O − ΔD) | 200 | DL: 100 | 35 | 10 | 0 |
| Including alternative paths with the same distance | 200 | DL: 100 | 57 | 8 | 0 |
| Including expert rules | 200 | DL: 79 RB: 21 | 70 | 0 | 0 |

### 4.2.3. 3D Scenario with Static Targets and Different Number of Obstacles

The next experiments consider a 3D scenario. As shown, when including another dimension, the results improve for the same number of obstacles (four obstacles in the first two rows) because the drones have more possibilities to reach the targets in a 3D environment. Figure 17 includes the principal results. The main parameters of the experimentation setup are the following:

- Reinforcement learning algorithm: PPO with a learning rate adjusted in preliminary experiments: 0.0003.
- 3D environment, including the Z component.
- Two agents (drones).
- Reward = 1 × T (number of targets reached) − 1 × O (number of obstacles hit) − ΔD (distance to target variation).
- Four static targets randomly distributed.
- The number of obstacles is variable, and they are randomly situated.

The main conclusions of these experiments are the following:

- As shown, when adding the third dimension (Z coordinate) with only four obstacles, the results on the percentage of episodes reaching all targets improves, reaching a very good value. This value (around 92%) is difficult to improve because, as targets and obstacles are situated randomly, there is always the possibility to have a target situated in the security zone of an obstacle and it cannot be reached.
- As predicted, when increasing the number of obstacles, the % of episodes reaching all targets decreases.
- An interesting result is that when increasing the number of obstacles, the drones do not hit obstacles, only one case when considering 20 obstacles.

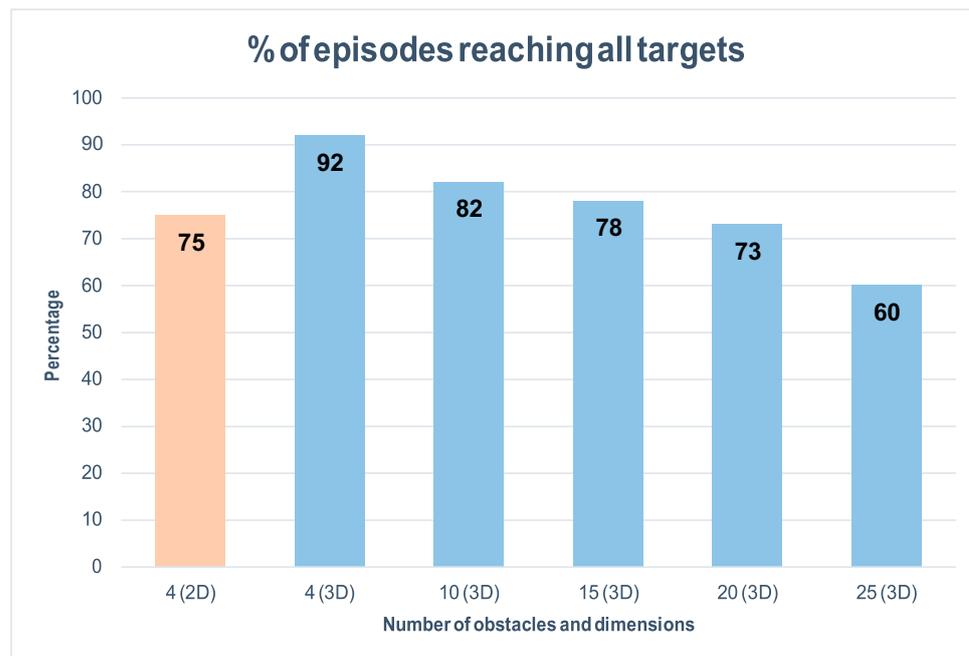

**Figure 17.** Main results when reaching targets with obstacles, considering a 3D scenario. Total used cycles in training is $8 \cdot 10^6$, and the maximum number of cycles per episode in testing is 200.

### 4.3. Task 2: Searching for Targets

This section presents the results obtained after evaluating the different strategies implemented for the second task: the agents (drones) must go over all the space searching for targets. This scenario includes several situations including or not including obstacles in the searching space and considering static or moving targets.

The main evaluation metrics considered in this study are the following:

- Quality metrics: to analyze the task success and the incidence of possible problems (like obstacles hits).
  - Percentage of episodes where all the targets were reached.
  - Percentage of episodes where at least one agent hit an obstacle.
  - Percentage of episodes where all agents hit an obstacle.
- Time-processing metrics: to analyze the drone movements required to complete the task.
  - Number of cycles used for training. To evaluate the amount of training required to train a good model.
  - Average number of cycles required to complete an episode. This number is divided into several numbers depending on the system state: average number of cycles per episode during testing, average number of cycles per episode during an exhaustive search, and average number of cycles per episode while searching based on the learnt RL model.

It is important to remark that not all metrics have been used in all the experiments. In these experiments, considering an exhaustive search, it is possible to discover all targets, so the most interesting performance metric is the reduction in the number of cycles/movements required to discover all the targets.

#### 4.3.1. Static and Moving Targets Without Obstacles

The first scenario considers a searching problem with static targets and without obstacles. The targets are distributed randomly in the scene organized in groups. The targets are distributed randomly in the scene organized in groups. The groups are also randomly situated in the scene. Every group occupies a small local area, and the targets are also randomly distributed inside the group area. Table 3 includes the main results obtained when searching targets without obstacles for a different number of groups. The main parameters of the experimentation setup are the following:

- Reinforcement learning algorithm: PPO with a learning rate adjusted in preliminary experiments: 0.0003.
- 2D environment.
- Two agents (drones).
- Reward when using RL $= 1 \times$ T (number of found targets) $- 1 \times$ O (number of obstacles hit) $- \Delta$D (distance to the first found target in the local zone).
- Four targets randomly distributed in several groups situated in zones with a size of 20% of the total width and height.

**Table 3.** Results regarding the different implementations for searching random static target without obstacles. Different number of groups.

| | Total Cycles in Training | Cycles per Episode in Testing: Initial Exhaustive Search | Cycles per Episode in Testing: RL Search | Cycles per Episode in Testing: Posterior Exhaustive Search | Total Number of Cycles in Testing | % of Episodes Reaching All Targets (Task Success) |
|---|---|---|---|---|---|---|
| Baseline system: only exhaustive search (1600 cycles is the maximum in the worst scenario) | $6 \times 10^6$ | 1232.4 | 0 | 0 | 1232.4 | 100 |
| Exhaustive and RL (1 group) | $6 \times 10^6$ | 660.9 | 39.5 | 128.7 | 829.1 | 100 |
| Exhaustive and RL (2 groups) | $6 \times 10^6$ | 638.1 | 62.4 | 245.3 | 945.8 | 100 |
| Exhaustive and RL (3 groups) | $6 \times 10^6$ | 623.0 | 88.1 | 351.2 | 1062.3 | 100 |
| Exhaustive and RL (4 groups) | $6 \times 10^6$ | 522.6 | 101.2 | 396.4 | 1020.2 | 100 |

From these results, the main conclusions are the following:

- The first aspect is that in all cases, it is possible to obtain a significant reduction (around 20%) in the total number of cycles in testing required to find all the targets compared to the baseline.
- When increasing the number of groups, the searching problem is more complicated and the number of cycles increases, except for four groups. In this case, as there are only four targets and four groups, the targets appear distributed along the total space, reducing the number of cycles in the initial exhaustive search (from 623.0 to 522.6).

The next experiments include moving targets: when any target is detected, the rest of the targets try to reach the bottom of the scene to escape from the scene. The main parameters of the experimentation setup are the following:

- Reinforcement learning algorithm: PPO with a learning rate adjusted in preliminary experiments: 0.0003.
- 2D environment.
- Two agents (drones).
- Reward when using RL $= 1 \times$ T (number of found targets) $- 1 \times$ O (number of obstacles hit) $- \Delta$D (distance to the first found target in the local zone)
- Four targets randomly distributed in several groups situated in zones with a size of 20% of the total width and height. When any target is detected, the rest start moving, trying to reach the bottom of the scene.

Table 4 includes the main results obtained in this situation.

**Table 4.** Results include movements of the targets trying to reach the bottom of the scene to escape from the scene.

| | Total Cycles Used in Training | Cycles per Episode in Testing: Initial Exhaustive Search | Cycles per Episode in Testing: RL Search | Cycles per Episode in Testing: Posterior Exhaustive Search | Total Number of Cycles in Testing | % of Episodes Reaching All Targets (Task Success) |
|---|---|---|---|---|---|---|
| Baseline system: only exhaustive search (1600 cycles is the maximum in the worst scenario) | $6 \times 10^6$ | 1222.4 | 0 | 0 | 1222.4 | 87 |
| Exhaustive and RL (1 group) | $6 \times 10^6$ | 648.6 | 43.4 | 241.1 | 933.1 | 89 |
| Exhaustive and RL (2 groups) | $6 \times 10^6$ | 569.6 | 73.1 | 370.8 | 1013.5 | 91 |
| Exhaustive and RL (3 groups) | $6 \times 10^6$ | 608.9 | 89.1 | 346.3 | 1044.3 | 90 |
| Exhaustive and RL (4 groups) | $6 \times 10^6$ | 568.1 | 98.4 | 311.4 | 977.9 | 90 |

From these results, the main conclusions are the following:

- The first aspect is that in all cases, it is possible to obtain a significant reduction (around 20%) in the total number of cycles required to find all the targets compared to the baseline. Similar results to the previous table.
- As the targets start moving when any of them are detected, there are cases where some targets reached the bottom of the scene, reducing the percentage of episodes reaching all targets. Values of around 90% were obtained in all cases, slightly lower in the baseline system because the number of required cycles is higher, so the targets have more time to reach the bottom of the scene.
- When increasing the number of groups, the searching problem is more complicated and the number of cycles increases, except for four groups. In this case, as there are only four targets and four groups, the targets appear distributed along the total space, reducing the number of cycles in the initial exhaustive search (from 608.9 to 568.1). Similar behavior was observed in the case of static targets.

### 4.3.2. Static Targets Including Obstacles in Task 2

Finally, this section describes the results when including obstacles in the searching mission with static obstacles (Table 5). The main parameters of the experimentation setup are the following:

- Reinforcement learning algorithm: PPO.
- Learning rate: 0.0003.
- 2D environment.
- Two agents (drones).
- Reward when using RL $= 1 \times T$ (number of found targets) $- 1 \times O$ (number of obstacles hit) $- \Delta D$ (distance to the first found target in the local zone).
- Four static targets randomly distributed in several groups situated in zones with a size of 20% of the total width and height.
- Five obstacles randomly distributed in the scene.

**Table 5.** Results regarding the different implementations for searching random static targets with obstacles. Different number of groups.

| | Total Cycles Used in Training | Cycles per Episode in Testing: Initial Exhaustive Search | Cycles per Episode in Testing: RL Search | Cycles per Episode in Testing: Posterior Exhaustive Search | Total Number of Cycles in Testing | % of Episodes Reaching All Targets |
|---|---|---|---|---|---|---|
| Baseline system: only exhaustive search (1600 cycles is the maximum in the worst scenario) | $6 \times 10^6$ | 1243.3 | 0 | 0 | 1243.3 | 95 |
| Exhaustive and RL (1 group) | $6 \times 10^6$ | 692.6 | 66.5 | 430.1 | 1189.2 | 94 |
| Exhaustive and RL (2 groups) | $6 \times 10^6$ | 565.4 | 92.4 | 510.3 | 1168.1 | 93 |
| Exhaustive and RL (3 groups) | $6 \times 10^6$ | 600.3 | 102.1 | 480.2 | 1182.6 | 94 |
| Exhaustive and RL (4 groups) | $6 \times 10^6$ | 583.6 | 111.3 | 604.4 | 1299.3 | 91 |

The next table includes the main results obtained in this situation.

The main conclusions from these results are the following:

- The first aspect is that the percentage of task completion does not reach 100% (95%) in the baseline system (exhaustive search). The reason is because there are targets that are situated in the margin area of an obstacle, and they cannot be discovered. When combining an exhaustive and RL search, similar percentages were obtained due to the same reason. Some cases where the agent gets stacked due to obstacles are observed, but these cases are less frequent.
- When considering less than four groups, a small reduction (5–10%) was obtained in the number of total cycles in testing required to find all the targets compared to the baseline. For the case of four groups, a worse result was obtained with a small increment in the number of cycles. When increasing the number of groups, the searching problem becomes more complicated, and the number of cycles increases.
- Comparing these results with those obtained without obstacles, there are smaller improvements in the number of cycles. This is due to the increment in task difficulty.

## 5. Conclusions

This paper describes the development and evaluation of hybrid artificial intelligence strategies for drone navigation in simulated environments. The hybrid AI combines deep learning models with rule-based strategies to generate the agent action based on the agent state. The system has a high level of configurability to adjust the scenario difficulty, including a different number of targets or obstacles. This tool incorporates explainable strategies for analyzing agent decisions and human interaction facilities for correcting or modifying agents' behaviors by a human operator.

From the experiments, the main conclusion is that hybrid AI, combining machine learning and rule-based engines, allows obtaining a very good compromise between performance and robustness. In the reaching scenario, the rule-based engine allows avoiding obstacles in a better way. For the searching scenarios, the exhaustive search based on expert rules has permitted the integration of RL models when the targets are located based on patterns that can be learnt using reinforcement learning.

As a general conclusion, it is possible to say that reinforcement learning is a very good strategy to learn policies for drone navigation. All the drones can include the same policy (same model), and this policy works well in different situations with different drones and target locations. Also, when considering moving targets, the drone policy can adapt itself to the new conditions without the need to retrain the model. This characteristic provides an RL-based solution with very good adaptability and scalability. In complex scenarios (like when considering obstacles), the RL model does not always learn all the possibilities, for example, for avoiding an obstacle in the path. In these situations, it is necessary to complement it with a rule-based module to have a very high task success rate.

The main limitation of these experiments is that hybrid AI strategies have been used in simulated environments including some simplifications. For example, the simulation assumes that one drone finds a target when both are in the same position. In a real scenario, this assumption cannot be true if the target is hidden.

For future work, two main directions have been considered. Firstly, more complex or realistic environments can be simulated, like including a different number of drones, targets, or obstacles, including weather conditions that affect the probability of detecting an obstacle, or wind models that modify the drone position without executing any movement. Secondly, new tasks can be implemented and simulated like crossing an environment without hitting any target or obstacle. This evaluation in more realistic environments would be necessary before onboarding these algorithms in real drones.

**Author Contributions:** Conceptualization, R.S.-S. and A.M.B.; methodology, R.S.-S., L.A., M.G.-M., D.C. and A.M.B.; software, R.S.-S., L.A., M.G.-M. and D.C.; validation, R.S.-S., M.G.-M., D.C. and A.M.B.; formal analysis, R.S.-S., M.G.-M., D.C. and A.M.B.; investigation, R.S.-S., L.A., D.C. and A.M.B.; resources, R.S.-S. and A.M.B.; data curation, R.S.-S., M.G.-M., D.C. and A.M.B.; writing—original draft preparation, R.S.-S. and M.G.-M.; writing—review and editing, R.S.-S., L.A., M.G.-M., D.C. and A.M.B.; visualization, R.S.-S., M.G.-M., D.C. and A.M.B.; supervision, R.S.-S. and A.M.B.; project administration, R.S.-S. and A.M.B.; funding acquisition, R.S.-S. and A.M.B. All authors have read and agreed to the published version of the manuscript.

**Funding:** This research was funded by the European Union's EDF programme under grant agreement number 101103386 (FaRADAI project).

**Institutional Review Board Statement:** Not applicable.

**Informed Consent Statement:** Not applicable.

**Data Availability Statement:** No new data were created or analyzed in this study. Data sharing is not applicable to this article.

**Acknowledgments:** Authors want to thank all the UPM colleagues in the FARADAI consortium for their comments and suggestions.

**Conflicts of Interest:** The authors declare no conflicts of interest.